\documentclass[runningheads]{llncs}

\usepackage{eccv}

\usepackage{eccvabbrv}

\usepackage{graphicx}
\usepackage{booktabs}

\usepackage[accsupp]{axessibility}  

\usepackage{hyperref}

\usepackage{orcidlink}

\newcommand{\bhline}[1]{\noalign{\hrule height #1}}
\usepackage{siunitx}

\begin{document}

\title{Reducing the Gap Between Spatial and Space-Filling Curve Neighbors in 3D Point Cloud Segmentation} 

\titlerunning{Abbreviated paper title}

\author{Konan Sasaki\inst{1}\orcidlink{0009-0009-7918-524X} \and
Junya Komori\inst{2} \and
Kazuhiro Hotta\inst{1}\orcidlink{0000-0002-5675-8713}}

\authorrunning{F.~Author et al.}

\institute{Meijo University, 1-501 Shiogamaguchi, Tempaku-ku, Nagoya, Aichi 468-8502, Japan \and
Asia Air Survey Co., Ltd., Shinyuri 21 Bldg., 1-2-2 Manpukuji, Asao-ku, Kawasaki, Kanagawa 215-0004, Japan \\
\email{sskkonan0628@gmail.com, jny.komori@ajiko.co.jp, kazuhotta@meijo-u.ac.jp} \\
}

\maketitle

\begin{abstract}
  As a lightweight and highly accurate method for nearest-neighbor search in 3D point clouds, approaches based on space-filling curves (SFC) have been proposed.
  However, due to the inherent properties of SFC, a gap arises between proximity relationships along the SFC and those in the original Euclidean space.
  In this study, we propose a loss function that reduces the influence of this gap.
  Specifically, we compute the distances between adjacent points along the SFC and train the model to optimize feature similarity between pairs of points that are distant from each other in the original 3D space. 
  We incorporate the proposed method into a high-performing 3D point cloud segmentation model and evaluate its effectiveness on three datasets.
  The results demonstrate that the proposed method outperforms conventional methods on both indoor and outdoor point cloud datasets.
\end{abstract}

\section{Introduction}
\label{sec:intro}

\begin{figure}[t]
  \centering
  \includegraphics[scale=0.25]{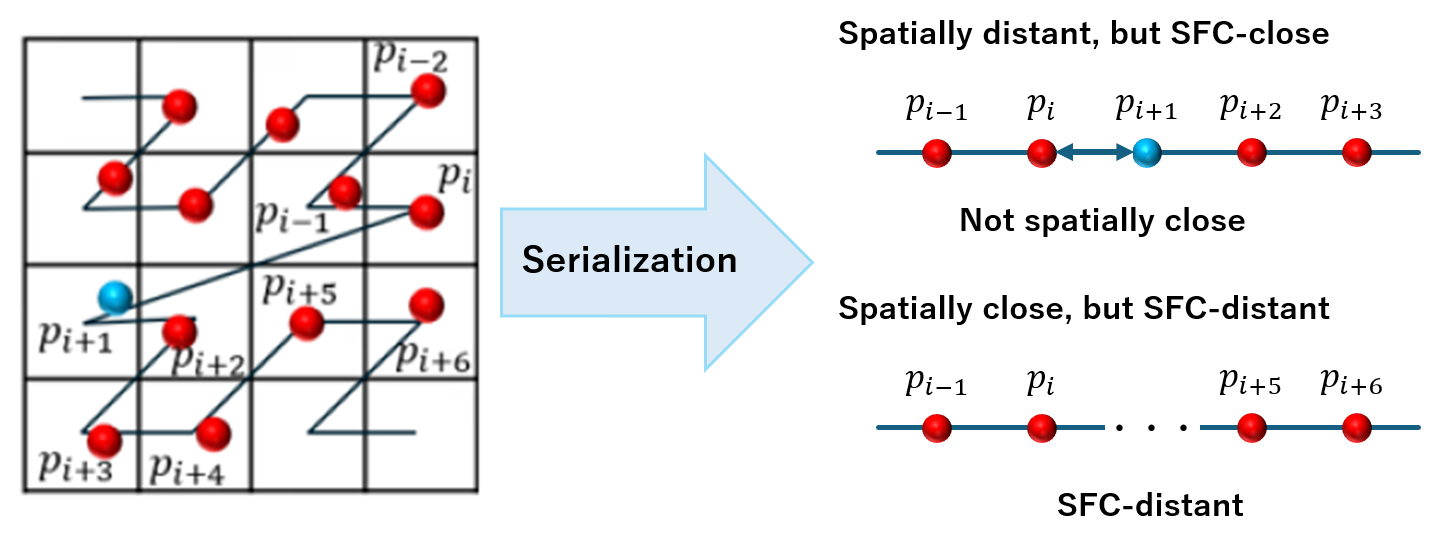}
   \caption{The gap between neighboring points in the original 3D space and those on the SFC. SFC-based serialization enables efficient neighborhood selection without performing the large number of distance calculations required by KNN. However, due to the characteristics of the SFC, points that are far apart in the original 3D space may be treated as neighboring points on the SFC (upper right).
   Conversely, points that are close to each other in the original 3D space may be treated as distant points on the SFC (lower right). }
    \label{fig:intro}
\end{figure}

\begin{figure*}[t]
  \centering
  \includegraphics[scale=0.25]{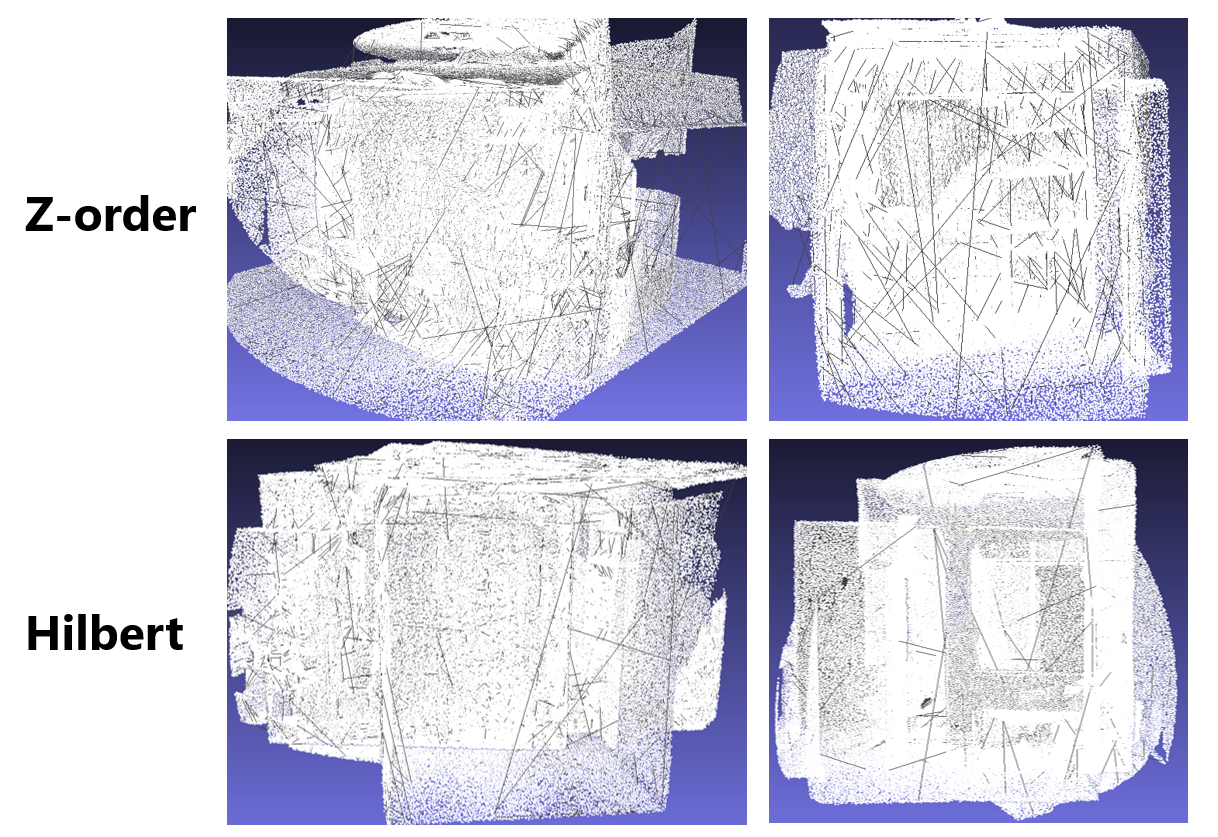}
   \caption{Visualization of adjacent points on the SFC that are spatially distant from each other. Point pairs whose distances exceed the threshold in \cref{eq:2} of the proposed method are visualized by connecting them with black lines. Since this point cloud is taken from S3DIS dataset during the training of Point Transformer V3, data augmentation has been applied.
   }
      \label{fig:intro2}
\end{figure*}

3D point cloud segmentation has a wide range of applications and is in high demand\cite{drive1,robot}. 
To achieve accurate point cloud segmentation, a model must correctly understand the detailed information contained in point clouds.
A key factor is the method used to search for neighboring points in order to capture local information in the point cloud.
Traditional and widely used neighborhood search methods are K-Nearest Neighbors (KNN) and radius search.
KNN computes the distances between a query point and all other points, and selects the K points with the smallest distances as its neighboring points.
Although this method is highly accurate, it must compute distances to a large number of points, resulting in high computational cost.
Radius search defines neighboring points as those located within a radius R from the query point.
However, since the number of neighboring points may vary depending on the point density, this approach can lead to unstable neighborhood search.
As described above, these search methods have issues in terms of computational cost and search accuracy.
To overcome these problems, Point Transformer V3\cite{ptv3}, a high-performing model for semantic segmentation of 3D point clouds, employs an SFC-based approach.
Specifically, the point cloud is ordered and rearranged using SFC, which preserves spatial locality to some extent, and the points within K positions before and after a query point along the SFC order are defined as its neighboring points.
This method can obtain neighboring points at low computational cost while preserving spatial information to some extent. 
However, as shown in \cref{fig:intro}, SFC may treat points that are far apart in the original 3D space as neighboring points.
As a result, the neighborhood relationships in the original 3D space and those on the SFC do not necessarily coincide, which leads to the problem that accurate neighboring points cannot be obtained.
To illustrate this in practice, \cref{fig:intro2} presents a visualization in which pairs of adjacent points on the SFC that are relatively distant in the original 3D space are connected by black lines.
The point cloud shown in the figure is taken from the training of Point Transformer V3 on S3DIS dataset, with data augmentation applied.
As can be seen from the figure, point pairs that are spatially distant and would not be considered neighbors by KNN or radius search are treated as adjacent points on the SFC. This indicates that neighborhood acquisition based on the SFC can incorrectly regard distant points as neighboring points.

In this study, we improve the segmentation accuracy of point clouds without increasing the number of model parameters or inference time by reducing the gap between neighboring points in the original 3D space and those on the SFC using a similarity-based loss function.
The proposed Distance-Aware Space-filling curves Loss (DAS Loss) is a loss function that strengthens the relationships 
between distant points by optimizing the feature similarity between points that are adjacent on the SFC but far apart in the original 3D space.
Specifically, we first set a threshold that is automatically determined for each point cloud scene. 
We then calculate the distances in the original 3D space between adjacent points on the SFC and apply contrastive loss to point pairs whose distance exceeds the threshold. 
This enables the gap between erroneous neighborhood relationships induced by the SFC and neighborhood relationships in the original 3D space to be mitigated in the feature space. 
Moreover, by focusing on spatially distant points during training, the model can efficiently learn global information in point clouds, thereby improving point cloud segmentation performance.
As a result, the model learns semantic information derived from feature similarity, rather than information derived from erroneous neighborhood relationships on the SFC, thereby reducing the impact of the gap between neighboring points in the original 3D space and those on the SFC.

In the experiments, we evaluated the effectiveness of the proposed method on the outdoor point cloud dataset DALES\cite{dales} and the indoor point cloud datasets: S3DIS\cite{s3dis} and ScanNet V2\cite{scannet}.
The proposed method achieved mIoU improvements of 1.62\% on DALES, 1.86\% on S3DIS, and 0.94\% on ScanNet V2.

This paper is organized as follows.
\cref{sec:related} discusses related work.
\cref{sec:method} describes our proposed method.
\cref{sec:exp} presents the results of the evaluation experiments.
Finally, \cref{sec:concl} concludes the paper and discusses future work.

\section{Related Work}
\label{sec:related}

\noindent \textbf{Point cloud understanding.}

Deep neural architectures for 3D point cloud understanding have conventionally been categorized into three main types according to how they represent and process point clouds: projection-based, voxel-based, and point-based methods.
Projection-based methods first project 3D points onto multiple 2D image planes and then employ 2D CNN backbones to extract features\cite{2dcnn1,2dcnn2,2dcnn3}.
Voxel-based methods convert point clouds into regular voxel grids, enabling the use of 3D convolutional operations\cite{voxel1,voxel2}.
Their computational efficiency has been further improved through sparse convolution techniques\cite{sparseconv1, sparseconv2}; however, these methods often suffer from limited scalability, particularly with respect to kernel size.
In contrast, point-based methods directly operate on raw point clouds\cite{pointbase1,pointnet,pointnetpp,pointnext}.
Recently, this category has increasingly adopted transformer-based architectures\cite{pct,transformerbase1,pt}.
Although these methods achieve strong performance, their scalability is often limited by the unstructured nature of point clouds.

Beyond these conventional categories, serialization-based methods\cite{serializedbase1,octformer,flatformer} have recently emerged as a promising approach to address the scalability limitations of point-based transformer architectures caused by the unstructured nature of point clouds.
These methods organize point clouds by sorting points according to predefined patterns, thereby converting unstructured and irregular point clouds into tractable sequences while partially preserving spatial locality.
OctFormer\cite{octformer} preserves point ordering through octree construction.
Although this design provides good scalability, its flexibility is still limited by the underlying octree structure.
In contrast, FlatFormer\cite{flatformer} adopts a window-based sorting strategy to group point pillars, resembling window partitioning.
However, this approach has limited scalability in terms of receptive field size and is primarily suited for pillar-based 3D object detection.
These pioneering studies laid the foundation for serialization-based point cloud processing.

Building upon this line of research, Point Transformer V3\cite{ptv3} was proposed to fully unlock the potential of point cloud serialization.
Point Transformer V3 serializes point clouds using the SFC, transforming irregular and unstructured point clouds into a more manageable one-dimensional sequence.
Based on this SFC ordering, neighboring points can be efficiently identified without relying on computationally expensive neighborhood search methods such as KNN.
Specifically, by defining the K points before and after a reference point on the SFC as neighboring points, the method avoids the computationally expensive process of calculating distances between points.
However, due to the characteristics of the SFC, this neighbor acquisition method has issues with accuracy.
As shown in \cref{fig:intro}, the ordering of the SFC does not necessarily connect neighboring regions and may also link distant regions.
Consequently, when neighboring points are defined according to the SFC order, spatially distant points may be treated as neighboring points, resulting in inaccurate neighborhood relationships.
These erroneous neighborhood relationships can help the model capture global relationships between distant points; however, they may also negatively affect its ability to understand the complex local structure of point clouds.

To address this issue, we propose applying a contrastive loss to the incorrectly defined SFC-based neighboring points.
This approach aims to reduce the gap, in the feature space, between SFC-based neighboring points and neighboring points in the original 3D space, while enhancing the global relationships within the point cloud.
As a result, the negative impact of SFC-based neighborhood selection on the model is mitigated, enabling more accurate point cloud segmentation.

\begin{figure*}[t]
  \centering
  \includegraphics[scale=0.49]{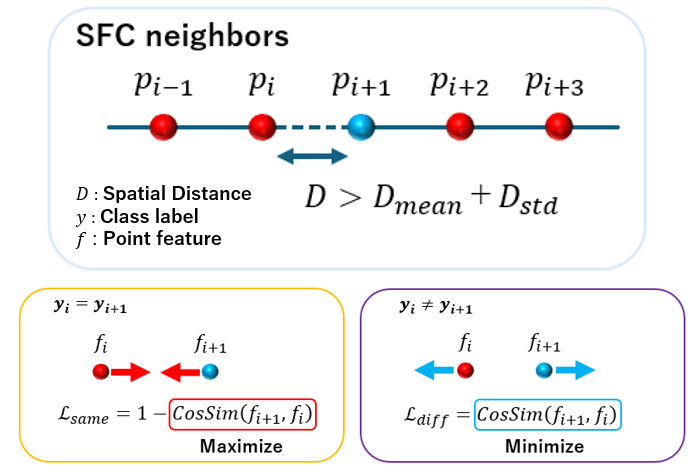}
   \caption{Overview of DAS Loss. The loss is adopted when the distance between adjacent points on the SFC exceeds the sum of the mean and standard deviation of all adjacent-point distances.
   If the class labels of the two points are identical, their feature representations are encouraged to be similar; otherwise, they are encouraged to be dissimilar.
   }
      \label{fig:method}
\end{figure*}

\section{Proposed Method}
\label{sec:method}

We describe the details of the proposed DAS Loss.
\cref{fig:method} shows the overview of the proposed method.
First, one SFC is randomly selected from four types of SFC: Z-order, Hilbert, Trans Z-order, and Trans Hilbert.
The four SFCs are the same as those used in Point Transformer V3\cite{ptv3}.
The point cloud is then reordered using the selected SFC.
After reordering, the distances between adjacent points on the SFC are calculated, and the loss function is applied to pairs of points whose distance exceeds a threshold.
For each point cloud scene, the threshold is computed as the sum of the mean distance $D_{mean}$ and the standard deviation $D_{std}$ of all distances between adjacent points, i.e., $D_{mean}+D_{std}$.
Since neighboring points on an SFC are often also neighboring points in the original 3D space, if the distance between adjacent points is smaller than the mean distance $D_{mean}$, they can be regarded as neighboring points in the original 3D space as well.
On the other hand, in order to identify points that are adjacent on the SFC but distant in the original 3D space, we use the value obtained by adding the standard deviation $D_{std}$ to the mean distance $D_{mean}$ as the threshold. 
This enables the threshold to be dynamically determined according to the distribution of point-to-point distances, allowing selective extraction of point pairs that are far apart in the original 3D space.
With these two parameters, an adaptive threshold can be determined for each point cloud scene.

Next, we explain the loss function applied to SFC-adjacent point pairs whose inter-point distance in the original 3D space exceeds this threshold.
When the class labels of two points are the same, the model is optimized so as to maximize the similarity between their features.
Conversely, when the class labels of the two points are different, the model is optimized to minimize the similarity between their features.
The final DAS Loss is obtained by averaging the loss functions calculated for these cases.
The proposed method described above can be formulated as shown in \cref{eq:1}–\cref{eq:5}.

\begin{align}
D_i &= \left\lVert p_{i+1} - p_i \right\rVert_2 \label{eq:1}\\
Mask_i &= 1 \quad \text{if }D_i > D_{\mathrm{mean}} + D_{\mathrm{std}}   \label{eq:2}\\
\mathcal{L}_{\mathrm{same}} ^{(j)}
&= 1 - \mathrm{CosSim}(f_{i+1}, f_i) 
\quad \text{if } Mask_i = 1 \ and \ y_{i+1}=y_i   \label{eq:3}\\
\mathcal{L}_{\mathrm{diff}}^{(k)} 
&= \mathrm{CosSim}(f_{i+1}, f_i) 
\quad \text{if } Mask_i = 1 \ and \ y_{i+1} \neq y_i   \label{eq:4}\\
\mathcal{L}_{\mathrm{DAS}}
&=
\frac{1}{2}
\left(
\frac{1}{N_{\mathrm{same}}}
\sum_{j=1}^{N_{\mathrm{same}}}
\mathcal{L}_{\mathrm{same}}^{(j)}
+
\frac{1}{N_{\mathrm{diff}}}
\sum_{k=1}^{N_{\mathrm{diff}}}
\mathcal{L}_{\mathrm{diff}}^{(k)}
\right)  \label{eq:5}
\end{align}

where $i = 0, 1, \ldots, N-2$. $N$ denotes the number of points in the point cloud, $p_i$ denotes the coordinates of the $i$-th point, $D_{mean}$ denotes the mean distance between all adjacent points, $D_{std}$ denotes the standard deviation of the distances between all adjacent points, $f_i$ denotes the feature of the $i$-th point, $y_i$ denotes the ground-truth class label of the $i$-th point, $\mathrm{CosSim}$ denotes the cosine similarity, and $N_{\mathrm{same}}$ and $N_{\mathrm{diff}}$ denote the number of valid point pairs satisfying the same-label and different-label conditions, respectively.
The final loss function is given by \cref{eq:6}.

\begin{align}
\mathcal{L}=\mathcal{L}_{CE}+\mathcal{L}_{lovasz}+\mathcal{L}_{DAS}  \label{eq:6}
\end{align}
where $\mathcal{L}_{CE}$ denotes the cross-entropy loss, $\mathcal{L}_{lovasz}$ denotes the Lovasz loss\cite{lovasz}, 
and $\mathcal{L}_{DAS}$ denotes DAS Loss.
The cross-entropy loss and Lovasz loss are the loss functions used in the conventional Point Transformer V3.
As described above, DAS Loss is a similarity-based loss function that leverages the SFC.
It aims to improve segmentation accuracy by reducing the adverse effects caused by the discrepancy between neighboring points on the SFC and those in the original 3D space.

\section{Experiments}
\label{sec:exp}

This section presents the experimental results used to verify the effectiveness of our method.
\cref{sec:det_exp} describes the experimental details.
\cref{sec:quant_exp} presents the quantitative evaluation, while  \cref{sec:qual_exp} provides the qualitative evaluation.
In addition, \cref{sec:abl_exp} presents an ablation study to further confirm the effectiveness of the proposed method.

\subsection{Details of the Experiments}
\label{sec:det_exp}
To evaluate the proposed method, we used DALES\cite{dales}, an outdoor point cloud dataset, as well as S3DIS\cite{s3dis} and ScanNet V2\cite{scannet}, which are indoor point cloud datasets.
DALES is a large-scale 3D outdoor point cloud dataset consisting of eight classes and acquired using an Aerial Laser Scanner (ALS).
The dataset consists of 29 tiles for training and 11 tiles for testing.
S3DIS consists of six large-scale indoor areas (Area 1–Area 6), 272 rooms, and 13 classes.
In this paper, we evaluated the proposed method using a commonly adopted evaluation protocol, in which Area 5 is used as the test set.
ScanNet V2 is a large-scale indoor dataset for segmentation and consists of 20 classes.
In this paper, we conducted the evaluation using 1,201 training point cloud scenes and 312 validation point cloud scenes.

\begin{table}[t]
    \centering
    \caption{Evaluation of the proposed method on DALES.}
    \label{tab : dales}

    \setlength{\tabcolsep}{3pt}

    \begin{tabular}{
        |c||
        S[table-format=2.2]@{\,$\pm$\,}S[table-format=1.2]
        S[table-format=2.2]@{\,$\pm$\,}S[table-format=1.2]
        |
    }
    \bhline{1.0pt}

    Class &
    \multicolumn{2}{c}{PTv3\cite{ptv3} (\%)} &
    \multicolumn{2}{c|}{Ours (\%)} \\

    \hline

    Ground
    & 95.59 & 0.13
    & \bfseries \textbf{95.97} & \textbf{0.08} \\

    Vegetation
    & 92.14 & 0.17
    & \bfseries \textbf{92.37} & \textbf{0.14} \\

    Cars
    & 83.77 & \textbf{0.06}
    & \bfseries \textbf{84.28} & 0.38 \\

    Trucks
    & 24.64 & 2.56
    & \bfseries \textbf{31.83} & \textbf{2.24} \\

    Power lines
    & 92.08 & \textbf{0.15}
    & \bfseries \textbf{93.18} & 0.61 \\

    Fences
    & 59.78 & \textbf{0.33}
    & \bfseries \textbf{60.70} & 0.66 \\

    Poles
    & 64.73 & 0.37
    & \bfseries \textbf{66.14} & \textbf{0.22} \\

    Buildings
    & 94.58 & 0.43
    & \bfseries \textbf{95.77} & \textbf{0.24} \\

    \hline

    mIoU
    & 75.91 & \textbf{0.41}
    & \bfseries \textbf{77.53} & 0.56 \\

    \bhline{1.0pt}

    \end{tabular}
\end{table}



In the experiments on DALES, the batch size was set to 4, the number of epochs to 2,500, and the initial learning rate to 2e-3.
For learning rate scheduling, the cosine learning rate schedule\cite{cosine}  was used.
Since random sampling was performed only once per point-cloud file in each epoch, the spatial coverage within a single epoch was relatively limited. 
Therefore, the model was trained for a large number of epochs to ensure sufficient coverage of the entire training set.
For data augmentation during training, we adopted the same techniques as those used in Point Transformer V3 (PTv3)\cite{ptv3}, including random rotation, random scaling, random flipping, and grid sampling.
In addition, to reduce VRAM usage in the experiments, the input point clouds were downsampled and spherical sampling was applied. During testing, points that were excluded from the computation due to downsampling were assigned the inference results of their nearest neighboring points, which were used as the final prediction results.

For S3DIS, we set the batch size to 8, the number of epochs to 1,000, and the initial learning rate to 6e-3, while for ScanNet V2, these values were set to 24, 800, and 6e-3, respectively.
In both experiments, the cosine learning rate schedule was used, and data augmentation followed the same settings as Point Transformer V3.
These experiments were conducted using two NVIDIA RTX A6000 GPUs, each with 48 GB of memory.
In addition, we also evaluated Point Transformer (PT)\cite{pt} and Point Transformer V2 (PTv2)\cite{ptv2}, which form the basis of Point Transformer V3.
For both Point Transformer and Point Transformer V2, the experiments on S3DIS were conducted for 1,000 epochs with a batch size of 8, whereas those on ScanNet V2 were conducted for 600 epochs with a batch size of 16.
All results for the methods evaluated in our experiments are reported as the mean and standard deviation over three independent runs.
For the other methods, we cited the results reported in their respective papers.
Since the original papers of Point Transformer and Point Transformer V2 report experiments only on indoor point clouds and single-object point clouds, applying these methods to outdoor point clouds is considered less practical due to their computational cost.
Therefore, for the experiments on DALES, we evaluate only Point Transformer V3, which is capable of efficiently processing point clouds.

As the evaluation metric, Intersection over Union (IoU) was used.
In the experiments described below, the evaluation was conducted using the IoU for each class and the mean IoU (mIoU), which is the average IoU over all classes.

\begin{table}[t]
\centering
\caption{Evaluation of the proposed method on S3DIS.}
\label{tab:s3dis}

\setlength{\tabcolsep}{3pt}

\begin{tabular}{
    |c||
    S[table-format=2.2]@{\,$\pm$\,}S[table-format=1.2]
    S[table-format=2.2]@{\,$\pm$\,}S[table-format=1.2]
    |
}
\bhline{1.0pt}

Class &
\multicolumn{2}{c}{PTv3\cite{ptv3} (\%)} &
\multicolumn{2}{c|}{Ours (\%)} \\

\hline

Ceiling
& 91.35 & 0.65
& \bfseries \textbf{93.69} & \textbf{0.50} \\

Floor
& 98.22 & 0.06
& \bfseries \textbf{98.40} & \textbf{0.04} \\

Wall
& 83.66 & 0.79
& \bfseries \textbf{84.12} & \textbf{0.26} \\

Beam
& 0.00 & 0.00
& 0.00 & 0.00 \\

Column
& \bfseries \textbf{34.21} & \textbf{1.40}
& 32.25 & 2.23 \\

Window
& 62.68 & 0.39
& \bfseries \textbf{62.80} & \textbf{0.16} \\

Door
& 72.88 & 1.86
& \bfseries \textbf{74.83} & \textbf{0.60} \\

Table
& \bfseries \textbf{83.73} & 0.79
& 82.45 & \textbf{0.37} \\

Chair
& 90.05 & 0.90
& \bfseries \textbf{92.65} & \textbf{0.62} \\

Sofa
& 67.71 & \textbf{0.97}
& \bfseries \textbf{79.20} & 2.68 \\

Bookcase
& 77.90 & 1.04
& \bfseries \textbf{79.58} & \textbf{0.44} \\

Board
& 83.55 & 2.77
& \bfseries \textbf{83.79} & \textbf{0.78} \\

Clutter
& 58.19 & 3.14
& \bfseries \textbf{64.73} & \textbf{1.41} \\

\hline

mIoU
& 69.56 & \textbf{0.27}
& \bfseries \textbf{71.42} & 0.42 \\

\bhline{1.0pt}

\end{tabular}
\end{table}

\begin{table}[t]
    \centering
    \caption{Evaluation of the proposed method on ScanNet V2.}
    \label{tab:scannet}

    \setlength{\tabcolsep}{3pt}

    \begin{tabular}{
        |c||
        S[table-format=2.2]@{\,$\pm$\,}S[table-format=1.2]
        S[table-format=2.2]@{\,$\pm$\,}S[table-format=1.2]
        |
    }
    \bhline{1.0pt}

    Class &
    \multicolumn{2}{c}{PTv3\cite{ptv3} (\%)} &
    \multicolumn{2}{c|}{Ours (\%)} \\

    \hline

    Wall
    & 87.38 & \textbf{0.32}
    & \bfseries \textbf{87.61} & 0.55 \\

    Floor
    & 95.64 & \textbf{0.05}
    & \textbf{95.88} & 0.16 \\

    Cabinet
    & 72.39 & \textbf{0.79}
    & \bfseries \textbf{73.12} & 1.20 \\

    Bed
    & 84.30 & \textbf{0.82}
    & \bfseries \textbf{87.09} & 1.69 \\

    Chair
    & 91.73 & 0.56
    & \bfseries \textbf{92.97} & \textbf{0.44} \\

    Sofa
    & 85.09 & \textbf{1.12}
    & \bfseries \textbf{86.46} & 1.44 \\

    Table
    & 76.40 & 1.38
    & \bfseries \textbf{78.61} & \textbf{0.49} \\

    Door
    & 72.36 & 0.72
    & \bfseries \textbf{73.04} & \textbf{0.67} \\

    Window
    & 70.63 & 1.08
    & \bfseries \textbf{72.36} & \textbf{0.87} \\

    Bookshelf
    & 82.88 & 1.88
    & \bfseries \textbf{83.43} & \textbf{0.97} \\

    Picture
    & \bfseries \textbf{38.96} & 1.58
    & 36.02 & \textbf{0.45} \\

    Counter
    & 70.97 & \textbf{0.32}
    & \bfseries \textbf{71.38} & 0.45 \\

    Desk
    & 69.18 & 2.22
    & \bfseries \textbf{71.42} & \textbf{1.35} \\

    Curtain
    & \bfseries \textbf{74.77} & \textbf{1.38}
    & 74.14 & 3.58 \\

    Refrigerator
    & 70.89 & 1.93
    & \bfseries \textbf{71.02} & \textbf{0.60} \\

    Shower curtain
    & 63.43 & \textbf{3.39}
    & \bfseries \textbf{68.72} & 3.58 \\

    Toilet
    & \bfseries \textbf{95.23} & \textbf{0.50}
    & 93.85 & 1.02 \\

    Sink
    & 69.56 & \textbf{1.08}
    & \bfseries \textbf{72.41} & 1.52 \\

    Bathtub
    & \bfseries 89.43 & 0.60
    & \textbf{89.56} & \textbf{0.42} \\

    Otherfurniture
    & 65.16 & 3.28
    & \bfseries \textbf{66.19} & \textbf{1.58} \\

    \hline

    mIoU
    & 76.32 & 0.25
    & \bfseries \textbf{77.26} & \textbf{0.14} \\

    \bhline{1.0pt}

    \end{tabular}
\end{table}

\subsection{Quantitative Evaluation}
\label{sec:quant_exp}
\cref{tab : dales} shows the IoU comparison between Point Transformer V3 and the proposed method on DALES.
Comparing Point Transformer V3 with the proposed method, our method achieved a 1.62\% improvement in mIoU.
Similarly, in the IoU comparison between Point Transformer-based models and the proposed method on S3DIS shown in \cref{tab:s3dis}, the proposed method achieved a 1.86\% improvement in mIoU.
In addition, in the comparison on ScanNet V2 shown in \cref{tab:scannet}, the proposed method achieved a 0.94\% improvement in mIoU.
These results confirm that the proposed DAS Loss is an effective loss function for both outdoor and indoor datasets.
This improvement can be attributed to the optimization of feature similarity between SFC-adjacent point pairs that are spatially distant in the original 3D space.
This optimization allows the model to learn appropriate semantic relationships between these points in the feature space.
Therefore, by introducing the loss function, the model can learn without relying solely on neighborhood relationships on the SFC.
As a result, among neighboring points on the SFC, the model becomes capable of properly handling points that are not neighboring points in the original 3D space.
This may reduce the influence of the gap between neighborhood relationships on the SFC and those in the original 3D space, leading to improved mIoU.

Regarding the class-wise IoU results, substantial improvements were observed for Cars and Trucks in DALES, as well as for Chair and Sofa in S3DIS and ScanNet V2, which are classes that are prone to misclassification due to their similarity.
This is considered to be because optimizing the similarity between features improved the separability between classes in the feature space, thereby suppressing misclassification among classes with similar shapes or appearances.
In addition, the IoU for Clutter, which contains objects of various sizes and types, was also significantly improved.
This may be because the optimization of feature similarity enabled the model to capture diverse intra-class characteristics, such as size, color, and shape.
In particular, greater improvements are expected for small objects that are easily misclassified because they are surrounded by points belonging to different classes.
Furthermore, since the model can learn relationships between spatially distant points, it is considered that its ability to semantically capture global information in point clouds also contributed to the improvement in IoU.
However, when comparing the degree of mIoU improvement across datasets, the improvement on ScanNet V2 was relatively small.
This may be attributed to the fact that ScanNet V2 contains noisier point cloud data than the other datasets.
Noise in the point clouds can make the extracted features less accurate.
As a result, the feature similarity optimization performed by the proposed method may not function effectively, which could explain the relatively small improvement in mIoU.
Although it is necessary to improve the robustness of the proposed method against noise in point clouds, the proposed method achieved a certain degree of improvement in mIoU.
This confirms the generality and effectiveness of the proposed method.

\cref{tab:othermethod} shows a performance comparison between the proposed method and various existing methods.
Compared with various existing methods, our proposed method achieved the highest mIoU and the lowest latency on ScanNet V2, while also showing the second-lowest memory usage, demonstrating highly competitive performance. On S3DIS, the proposed method achieved the second-highest mIoU among the compared existing methods.
Although Swin3D\cite{swin3d} achieves the highest mIoU on S3DIS, it requires significantly higher latency and memory usage than our proposed method.
Therefore, when considering both mIoU and computational cost, our proposed method can be regarded as achieving the best overall performance on S3DIS as well.
A major advantage of the proposed method is that we can improve segmentation accuracy without modifying the model architecture or increasing latency and memory usage during inference, because it is introduced only as a loss function during training.
These results confirm that, across the two datasets, our proposed method achieved the best overall performance among all compared methods.

\begin{table}[t]
   \centering
   \caption{Performance comparison between existing methods and the proposed method on S3DIS and ScanNet V2. Bold numbers indicate the best performance among the compared methods, and underlined numbers indicate the second-best performance. Latency and memory are measured during inference on ScanNet V2. Results for PT and PTv2 are averaged over three independent runs conducted by us.
   }
   {
   \begin{tabular}{|l||cccc|} \bhline{1.0pt} 
       Method & ScanNet V2\cite{scannet} (\%)& S3DIS\cite{s3dis} (\%) & Latency & VRAM\\ 
   \bhline{1.0pt} 
   \hline
       BAAF-Net\cite{baafnet}  & - & 65.4 & - & - \\
       MinkUNet\cite{sparseconv1} & 72.2 & 65.4 & 126ms & \textbf{4.7G}  \\
       OctFormer\cite{octformer} & 75.7 & - & \underline{112ms} & 12.5G \\
       PointNeXt\cite{pointnext} & 71.5 & 70.5 & - & - \\
       PanopticNDT\cite{panoNDT} & 68.4 & - & - & - \\
       SegGCN\cite{seggcn} & 58.9 & 63.6 & - & - \\
       Swin3D\cite{swin3d} & \underline{76.4} & \textbf{72.5} & 593ms & 8.8G \\
       PT\cite{pt} & 75.3 & 67.8 & 1442ms & 7.1G \\
       PTv2\cite{ptv2} & 74.7 & 69.1 & 235ms & 18.2G  \\
       PTv3\cite{ptv3} & 76.3 & 69.6 & \textbf{85ms} & \underline{5.2G}  \\
       Ours & \textbf{77.3} & \underline{71.4} & \textbf{85ms} & \underline{5.2G}\\    
   \bhline{1.0pt} 
   \end{tabular}
   }
   \label{tab:othermethod}
\end{table}

\begin{figure*}[t]
  \centering
  \includegraphics[scale=0.28]{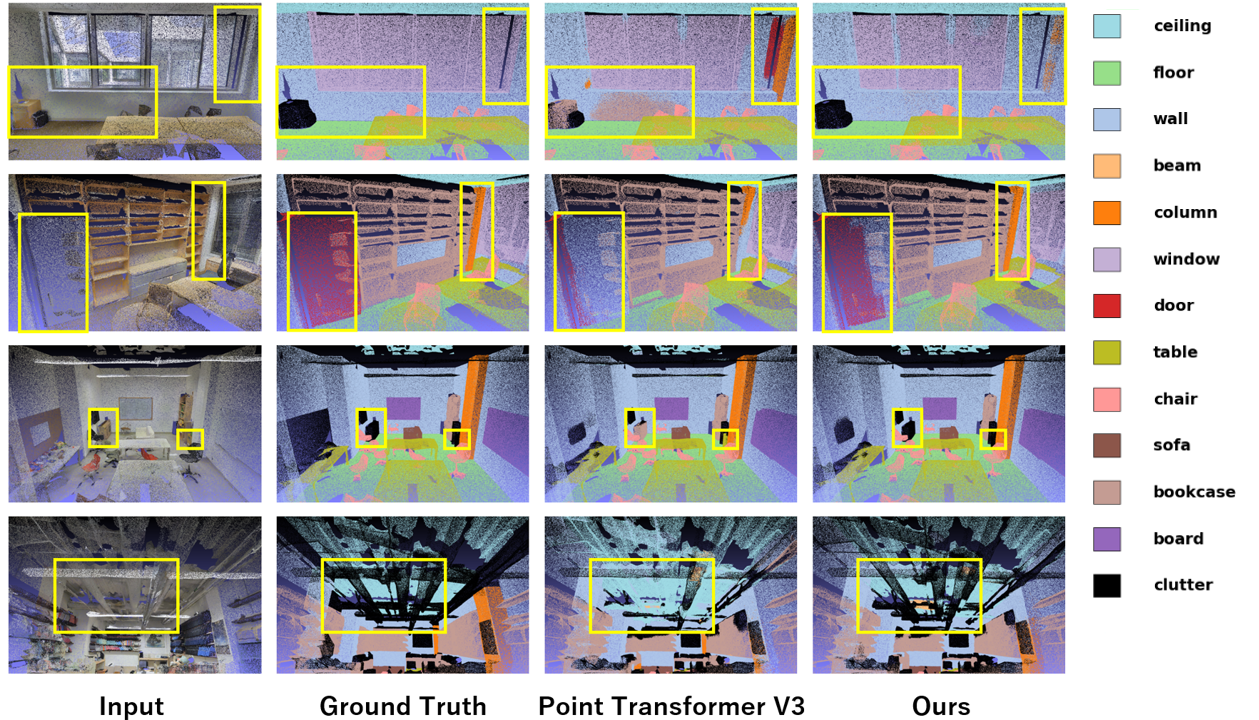}
   \caption{Qualitative evaluation of the proposed method on S3DIS. The input point cloud is visualized with original colors in the 1st column. The areas where improvements were observed compared to the conventional model are highlighted with yellow boxes in the figure.
   }
      \label{fig:vis}
\end{figure*}

\subsection{Qualitative Evaluation}
\label{sec:qual_exp}

\cref{fig:vis} shows the segmentation results on S3DIS.
In Point Transformer V3, partial misclassification occurs within objects, whereas the proposed method produces more consistent and improved prediction labels within the same object.
This result is considered to be due to the similarity constraint imposed by the loss function, which enables the model to learn semantic consistency among features.
In addition, since the column class is adjacent to the wall class, Point Transformer V3 produces misclassifications near object boundaries.
In contrast, the proposed method enables the model to understand the semantic relationships between points, allowing it to accurately identify points without being affected by neighboring points belonging to different classes.
For the same reason, the proposed method is also considered to reduce misclassification of small objects.
These results qualitatively confirm the effectiveness of the proposed method.

\subsection{Ablation Study}
\label{sec:abl_exp}

In this section, we describe an ablation study conducted to further verify the effectiveness of the proposed DAS Loss.
First, \cref{tab:ablate1} shows the experimental results for evaluating each component of DAS Loss, namely $\mathcal{L}_{same}$, $\mathcal{L}_{diff}$, and $\mathcal{L}_{DAS}$.
These results correspond to the cases where only $\mathcal{L}_{same}$, only $\mathcal{L}_{diff}$, and $\mathcal{L}_{DAS}$ were introduced into Point Transformer V3 as the loss function.
From these results, it can be seen that, for S3DIS and ScanNet V2, the best performance is achieved by using both $\mathcal{L}_{same}$, which maximizes the feature similarity between distant points belonging to the same class, and $\mathcal{L}_{diff}$, which minimizes the feature similarity between distant points belonging to different classes. 
In contrast, for DALES, the model using only $\mathcal{L}_{diff}$ achieved the highest mIoU.
This may be because outdoor point clouds tend to have more complex and diverse shapes and local structures within the same class than indoor point clouds, resulting in a wider feature distribution.
Therefore, uniformly increasing the feature similarity between distant points of the same class may suppress useful intra-class feature differences.
In point cloud segmentation, such feature differences can provide useful information when the model learns diverse point cloud characteristics, as suggested by point cloud feature routing using MoE \cite{pointmoe}.
On the other hand, $\mathcal{L}_{diff}$, which promotes feature separation between different classes, can improve discriminability while preserving such broad feature distributions, leading to higher accuracy.
However, when only $\mathcal{L}_{diff}$ is used for S3DIS and ScanNet V2, the improvement in mIoU is considerably smaller than when $\mathcal{L}_{DAS}$ is used.
Therefore, from an overall perspective, $\mathcal{L}_{DAS}$, which incorporates all components, is considered to be the most effective.

\begin{table}[t]
   \centering
   \caption{Evaluation experiments for each component of DAS Loss. 
   For $\mathcal{L}_{same}$ and $\mathcal{L}_{diff}$, the averaging operation in \cref{eq:5} is not applied. Instead, each loss is directly introduced into Point Transformer V3 as the final loss function.
}{
   \begin{tabular}{|l||ccc|} \bhline{1.0pt} 
   Dataset & $\mathcal{L}_{same}$ (\%)& $\mathcal{L}_{diff}$ (\%) & $\mathcal{L}_{DAS}$ (\%)\\
   \hline
       DALES\cite{dales} & 77.16 & \textbf{77.77} & 77.53 \\
       S3DIS\cite{s3dis} & 71.20 & 70.50 & \textbf{71.42} \\
       ScanNet V2\cite{scannet} & 76.63 & 76.80 & \textbf{77.26} \\
   \bhline{1.0pt} 
   \end{tabular}
   }
   \label{tab:ablate1}
\end{table}


\begin{table}[t]
   \centering
   \caption{Evaluation of the DAS Loss threshold. DAS Loss is applied to all point pairs when the threshold is 0.0, and only to spatially distant point pairs when the threshold is $D_{mean}+D_{std}$.
   Changing the threshold and inequality condition enables DAS Loss to target either spatially close or distant point pairs.}
   {
   \begin{tabular}{|c||ccc|}
   \bhline{1.0pt}
   Threshold & DALES\cite{dales} (\%) & S3DIS\cite{s3dis} (\%) & ScanNet V2\cite{scannet} (\%) \\
   \hline
   $D_i > 0.0$ & 76.20 & 70.41 & 76.45 \\
   $D_i < D_{\mathrm{mean}} - D_{\mathrm{std}}$ & 76.48 & 70.07 & 76.79 \\
   $D_i > D_{\mathrm{mean}} + D_{\mathrm{std}}$ & \textbf{77.53} & \textbf{71.42} & \textbf{77.26} \\
   \bhline{1.0pt}
   \end{tabular}
   }
   \label{tab:ablate2}
\end{table}

\cref{tab:ablate2} shows the experimental results for evaluating the effectiveness of the threshold in DAS Loss.
For each dataset, we compared three threshold settings. 
In the first setting, the threshold was set to 0.0. Namely, DAS Loss was applied to all point pairs. 
In the second setting, the inequality condition used in the threshold of \cref{eq:2} was reversed so that DAS Loss was applied to point pairs whose distances between SFC-based neighboring points are spatially small.
The third setting corresponds to our proposed method, in which DAS Loss is applied only to spatially distant point pairs.
When DAS Loss was applied to spatially close point pairs, no substantial improvement in accuracy was observed on DALES or S3DIS.
This indicates that DAS Loss is more effective when applied to spatially distant point pairs rather than spatially close ones.
These results suggest that optimizing the features of spatially distant point pairs contributes more effectively to accurate point cloud segmentation than optimizing the features of spatially close point pairs.
This may be because spatially close point pairs often belong to the same class, and therefore feature optimization between point pairs belonging to different classes is insufficiently performed.
Object boundary regions, where errors are likely to occur, often consist of point pairs from different classes.
Therefore, when applying a loss function to optimize the features of spatially close point pairs, it is important to include a sufficient number of point pairs belonging to different classes in the feature optimization process.
Specifically, in DALES, points involved in adjacent point pairs whose inter-point distances exceeded the threshold used in DAS Loss accounted for approximately 6\% of the entire point cloud. 
Therefore, we consider that the proposed loss function can be effective if the inter-point distances of a few percent of point pairs consisting of points from different classes exceed the threshold.
Although this proportion is only a few percent, outdoor point clouds are typically large-scale and contain a vast number of points. 
Consequently, the loss function is applied to a sufficiently large number of points.

For ScanNet V2, the highest mIoU was also achieved when feature optimization was applied to spatially distant point pairs.
However, unlike the results for DALES and S3DIS, ScanNet V2 showed a certain improvement in mIoU even when the features of spatially close point pairs were optimized.
This may be because the negative effects of noise, which is frequently observed in ScanNet V2, were mitigated by the feature optimization introduced by the proposed loss function.
Noise in point clouds can make local feature representations unstable.
By optimizing the features of spatially close point pairs, the consistency of features within local neighborhoods may be improved, thereby suppressing the influence of noise.
This suggests that, for point clouds with substantial noise, optimizing the features of spatially close point pairs using the loss function can also be effective to some extent.

Considering the overall results across the three datasets, applying DAS Loss to spatially distant point pairs, which optimizes their feature representations, is considered to be the most effective strategy for point cloud segmentation.
Next, compared with applying DAS Loss to all point pairs by setting the threshold to 0.0, applying DAS Loss to SFC-based neighboring point pairs that are spatially distant achieves higher mIoU.
This indicates that it is more effective to apply the loss function by focusing on spatially distant points rather than applying it to all points.
This may be because the threshold allows DAS Loss to be applied to global relationships between distant points, enabling the model to focus on global information that is important for understanding point clouds.
In addition, rather than optimizing the features of all points, focusing on reducing the gap in the feature space between SFC-based neighboring points and neighboring points in the original 3D space is considered to be important for improving mIoU.
These results confirm the effectiveness of the threshold in our proposed method.
Furthermore, they demonstrate that the proposed loss function has high generality and an appropriate formulation for point cloud segmentation.

\begin{table}[t]
   \centering
   \caption{Evaluation of different loss application conditions.
   "Close-$\mathcal{L}$" applies $\mathcal{L}$ to SFC-based neighboring point pairs that are spatially close, whereas "Distant-$\mathcal{L}$" applies $\mathcal{L}$ to SFC-based neighboring point pairs that are spatially distant. In the "Close" setting, we use the threshold $D_i < D_{\mathrm{mean}} - D_{\mathrm{std}}$, whereas in the "Distant" setting, we use $D_i > D_{\mathrm{mean}} + D_{\mathrm{std}}$.}
   {
   \begin{tabular}{|l|ccc|} \bhline{1.0pt}
   Loss Application & DALES\cite{dales} (\%) & S3DIS\cite{s3dis} (\%) & ScanNet V2\cite{scannet} (\%) \\
   \hline
   Close-$\mathcal{L}_{diff}$ \& Distant-$\mathcal{L}_{same}$ & 77.35 & 69.43 & 77.16 \\
   Close-$\mathcal{L}_{same}$ \& Distant-$\mathcal{L}_{diff}$ & 77.13 & 71.36 & \textbf{77.44} \\
   Distant-$\mathcal{L}_{same}$ \& Distant-$\mathcal{L}_{diff}$ & \textbf{77.53} & \textbf{71.42} & 77.26 \\
   \bhline{1.0pt} 
   \end{tabular}
   }
   \label{tab:ablate4}
\end{table}

Finally, \cref{tab:ablate4} presents an ablation study on the extensibility of the proposed method. 
The extensibility of the loss function is evaluated by changing the conditions under which the loss function is applied. 
In DALES and S3DIS, the highest mIoU is achieved when contrastive loss is applied by focusing on spatially distant points. 
However, for ScanNet V2, the highest mIoU is obtained when $\mathcal{L}_{same}$ is applied to SFC-adjacent points that are spatially close, while $\mathcal{L}_{diff}$ is applied to spatially distant points. 
This is considered to be because the influence of noise, which is abundant in ScanNet V2, is suppressed by applying $\mathcal{L}_{same}$ to local regions.
By maximizing the feature similarity in local regions, the influence of noise is minimized and feature consistency is enhanced, leading to more accurate point cloud segmentation. 
In addition, applying $\mathcal{L}_{diff}$ to spatially distant points improves inter-class discriminability, which is also considered to contribute to suppressing misclassification caused by noise. 
These results indicate that DAS Loss can achieve higher performance by customizing the loss function according to various purposes. 
Furthermore, by further subdividing the conditions under which the loss function is applied, DAS Loss has the potential to achieve even higher performance.

\section{Conclusion}
\label{sec:concl}

In this study, we proposed DAS Loss, a loss function that encourages the model to learn semantic relationships based on feature similarity rather than incorrect neighborhood relationships on the SFC.
The proposed loss reduces the influence of the discrepancy between neighboring points in the original 3D space and those on the SFC.
Consequently, it makes Point Transformer V3 a more accurate segmentation model on both outdoor and indoor point cloud segmentation tasks.
A key advantage of the proposed method is that the improvement is achieved solely by introducing the loss function. Therefore, it does not increase the number of model parameters or inference time.
One important issue to be addressed in this study is robustness to noise. 
The improvement in mIoU on ScanNet V2, which contains a large amount of noise, was smaller than the improvements observed on the other two datasets. 
This result indicates that noise may reduce the reliability of both feature similarity computation and the selection of point pairs to which DAS Loss is applied.
To overcome this limitation, we will focus on introducing noise-robust mechanisms into DAS Loss. 
Specifically, we plan to investigate adaptive thresholding strategies that are less affected by noisy distance distributions, and robust similarity measures for feature comparison.
\bibliographystyle{splncs04}
\bibliography{main}
\end{document}